\documentclass[10pt,journal,compsoc]{IEEEtran}
\renewcommand{\baselinestretch}{0.98}
\ifCLASSOPTIONcompsoc
  \usepackage[nocompress]{cite}
\else
  \usepackage{cite}
\fi

\usepackage{graphicx}
\usepackage{comment}
\usepackage{amsmath,amssymb} % define this before the line numbering.
\usepackage{color}

\usepackage{multirow}
\usepackage{multicol}
\usepackage{floatrow}
\usepackage[noend]{algpseudocode}
\usepackage{algorithmicx,algorithm}
\usepackage{bbding}

\usepackage{tikz,xcolor,hyperref}% Make Orcid icon
\usepackage{subfigure}
\usepackage{booktabs}
\usepackage{makecell}
\usepackage{xspace}

\definecolor{lime}{HTML}{A6CE39}
\DeclareRobustCommand{\orcidicon}{%
    \begin{tikzpicture}
    \draw[lime, fill=lime] (0,0) 
    circle [radius=0.16] 
    node[white] {{\fontfamily{qag}\selectfont \tiny ID}};    \draw[white, fill=white] (-0.0625,0.095) 
    circle [radius=0.007];    \end{tikzpicture}
    \hspace{-2mm}}
\foreach \x in {A, ..., Z}{%
    \expandafter\xdef\csname orcid\x\endcsname{\noexpand\href{https://orcid.org/\csname orcidauthor\x\endcsname}{\noexpand\orcidicon}}
    }

\begin{document}
\title{Audit to Forget: A Unified Method to Revoke Patients' Private Data in Intelligent Healthcare}

% \author{Michael~Shell,~\IEEEmembership{Member,~IEEE,}
%         John~Doe,~\IEEEmembership{Fellow,~OSA,}
%         and~Jane~Doe,~\IEEEmembership{Life~Fellow,~IEEE}% <-this % stops a space
% \author{Chong~Mou, Jian~Zhang
\author{Juexiao~Zhou$^{1,2,\#}$, Haoyang Li$^{1,2,\#}$, Xingyu Liao$^{1,2}$, Bin Zhang$^{1,2}$, Wenjia He$^{1,2}$, Zhongxiao Li$^{1,2}$, Longxi Zhou$^{1,2}$, Xin~Gao$^{1,2,*}$
\thanks{
$^1$Computer Science Program, Computer, Electrical and Mathematical Sciences and Engineering Division, King Abdullah University of Science and Technology (KAUST), Thuwal 23955-6900, Kingdom of Saudi Arabia\\
$^2$Computational Bioscience Research Center, King Abdullah University of Science and Technology, Thuwal 23955-6900, Kingdom of Saudi Arabia\\
$^*$Corresponding author\\
$^\#$Equal contribution
}}


\markboth{AFS}{}%
\IEEEtitleabstractindextext{%
\begin{abstract}
Revoking personal private data is one of the basic human rights, which has already been sheltered by several privacy-preserving laws in many countries. However, with the development of data science, machine learning and deep learning techniques, this right is usually neglected or violated as more and more patients' data are being collected and used for model training, especially in intelligent healthcare, thus making intelligent healthcare a sector where technology must meet the law, regulations, and privacy principles to ensure that the innovation is for the common good. In order to secure patients' right to be forgotten, we proposed a novel solution by using auditing to guide the forgetting process, where auditing means determining whether a dataset has been used to train the model and forgetting requires the information of a query dataset to be forgotten from the target model. We unified these two tasks by introducing a new approach called knowledge purification. To implement our solution, we developed AFS, a unified open-source software, which is able to evaluate and revoke patients' private data from pre-trained deep learning models. We demonstrated the generality of AFS by applying it to four tasks on different datasets with various data sizes and architectures of deep learning networks. The software is publicly available at \url{https://github.com/JoshuaChou2018/AFS}. 
\end{abstract}

}

\maketitle

\IEEEdisplaynontitleabstractindextext

% \fi
\IEEEpeerreviewmaketitle

\IEEEraisesectionheading{\section{Introduction}\label{sec:introduction}}
\IEEEPARstart{R}{evoking} personal private data is one of the basic human rights, which has already been sheltered by privacy-preserving regulations like The General Data Protection Regulation (GDPR)\cite{voigt2017eu}, The Health Insurance Portability and Accountability Act of 1996 (HIPAA)\cite{act1996health}, and the California Consumer Privacy Act \cite{pardau2018california} since $20^{th}$ century. With those regulations, users are allowed to request the deletion of their own data for privacy concerns and to secure their own `right to be forgotten'. However, with the development of data science, machine learning (ML) and deep learning (DL) techniques, this basic right is usually neglected or violated. For example, it has been observed that patients' genetic markers were leaked from ML methods for genetic data processing\cite{wang2009learning, fredrikson2014privacy} while the patients were unaware of that. When users realize the existence of such risks, they may request their own data to be deleted to protect their privacy\cite{cao2015towards}. Meanwhile, those aforementioned regulations will force involved third parties to take actions immediately. According to the requirements of those regulations, not only the previously authorized data by individuals need to be deleted immediately from hosts’ storage systems but also the associated information should be removed from DL models trained with those data, because DL models could memorize sensitive information of training data and thus expose individual’s privacy under risk\cite{fredrikson2015model, song2017machine, ganju2018property, carlini2019secret, zhou2022ppml}. 

Nowadays, healthcare is one of the most promising areas for the deployment of artificial intelligent (AI) systems as so-called intelligent healthcare. ML and DL-based computer-aided diagnosis (CAD) systems in intelligent healthcare accelerate the diagnosis of various diseases and achieve even better results than doctors, such as tumour detection\cite{mckinney2020international, ardila2019end}, retinal fundus imaging\cite{poplin2018prediction}, detection and segmentation of COVID-19 lung infections\cite{zhou2020rapid,zhou2022interpretable} and so on. However, as more and more patients' data are being collected and used for model training in intelligent healthcare, their privacy is exposed to high risk. Therefore, intelligent healthcare is a sector where technology must meet the law, regulations, and privacy principles to ensure that the innovation is for the common good\cite{bartoletti2019ai}. To obey those privacy-preserving regulations, methods to revoke personal private data from pre-trained DL models are necessary. 

Deleting the stored personal data is simple, whereas forgetting individuals’ private information from pre-trained DL models could be difficult as we could not fully measure the contribution of individual data on the training process of DL models due to the stochasticity of training\cite{bourtoule2021machine}. Besides, due to the incremental nature of training, the model update brought by one sample would affect the model performance on samples followed, thus making it difficult to unlearn\cite{bourtoule2021machine}. Finally, catastrophic unlearning might happen and the unlearned model will perform worse than the model retrained on the remaining dataset\cite{nguyen2020variational}. 

In general, the process to forget data from a pre-trained DL model could be divided into two steps. Firstly, the unlearning process (forgetting) is performed on a given pre-trained DL model to forget the target data with different techniques and a new DL model will be generated. Secondly, an evaluation of the new model (auditing) against different metrics will be performed to prove that the model has forgotten the target data. These two processes should be repeated until the new model passes the evaluation. In simple terms, there are two commonly acknowledged sub-tasks, which could also be stated in the reverse order: auditing and forgetting, as a two-player game. Auditing requires auditors to precisely evaluate whether the data of certain patients were used to train the target DL model. Once the data of certain patients is confirmed to be used to train the target DL model by auditing, forgetting requires the removal of learnt information of certain patients’ data from the target DL model, which is also called machine unlearning, while auditing could act as the verification of machine unlearning \cite{bourtoule2021machine}

In order to achieve forgetting, existing unlearning methods could be classified into three major classes, including model-agnostic methods, model-intrinsic methods and data-driven methods\cite{nguyen2022survey}. Model-agnostic methods refer to algorithms or frameworks that can be used for different DL models, including differential privacy\cite{gupta2021adaptive, bourtoule2021machine, thudi2022unrolling}, certified removal\cite{guo2019certified, golatkar2020eternal, neel2021descent}, statistical query learning\cite{cao2015towards}, decremental learning\cite{ginart2019making}, knowledge adaptation\cite{chundawat2022can, kim2022efficient} and parameter sampling\cite{nguyen2022markov}. Model-intrinsic approaches are those methods designed for specific types of models, such as for softmax classifiers\cite{baumhauer2022machine}, linear models\cite{izzo2021approximate}, tree-based models\cite{schelter2021hedgecut} and Bayesian models\cite{nguyen2020variational}. Data-driven approaches focus on the data itself, including data partitioning\cite{bourtoule2021machine}, data augmentation\cite{shan2020protecting, tarun2021fast, huang2021unlearnable} and other unlearning strategies based on data influence\cite{peste2021ssse}. All methods have their specific application scenarios and limitations. Among the three methods, model-agnostic methods might have the strongest application prospects, as they can be applied to different models. Still, more mechanisms and theoretical concepts are being proposed to explore different solutions to the forgetting task but few of them focused on the application in real-world intelligent healthcare. 

When forgetting is accomplished, auditing is the next necessary step to verify it. Different metrics have been proposed to audit the membership of the query dataset, including accuracy, completeness\cite{cao2015towards}, unlearn time, relearn time, retrain time, layer-wise distance, activation distance, JS-divergence, membership inference\cite{liu2020have, huang2021mathsf}, ZRF score\cite{chundawat2022can}, epistemic uncertainty\cite{hullermeier2021aleatoric} and model inversion attack\cite{fredrikson2015model}. In recent studies, membership inference-based metrics were frequently utilized to determine whether or not any information about the samples to be forgotten was retained in the model in intelligent healthcare\cite{huang2021mathsf}. A black-box setting was shared by the membership inference attack (MIA) to calculate the probability of a single datapoint being a member of the training dataset $D$. Based on this individual level MIA, Liu et al.\cite{liu2020have} and Yangsibo et al.\cite{huang2021mathsf} focused on a more challenging task: audit the membership of a set of data points. The ensembled membership auditing (EMA)\cite{huang2021mathsf} was proposed as the state-of-the-art method to verify whether a query dataset is memorized by a pre-trained DL model, which is also a benchmark metric in machine unlearning. However, due to the black box property of DL models, efficient and accurate auditing is still challenging and an under-studied topic. Moreover, researchers have tended to treat auditing and forgetting as separate tasks, ignoring the fact that the two can be linked up associatively to work as a self-consistent mechanism.

Here, we proposed a novel solution by using auditing to guide the forgetting process in a negative feedback manner. We unified the two tasks by introducing knowledge purification (KP), a new approach to selectively transfer the needed knowledge to forget the target information instead of simply transferring all information like knowledge distillation (KD)\cite{hinton2015distilling}. On the basis of KP, we have developed a user-friendly and open-source method called AFS, which can be easily used to revoke patients' private data from DL models in intelligent healthcare. To demonstrate the generality of AFS, we applied it to four tasks based on four datasets, including the MNIST dataset, the PathMNIST dataset, the COVIDx dataset and the ASD dataset, with different data sizes and various architectures of deep learning networks. Our results demonstrate the usability of AFS and its application potential in real-world intelligent healthcare.

\begin{figure*}[t]
    \centering
    \includegraphics[width=.95\linewidth]{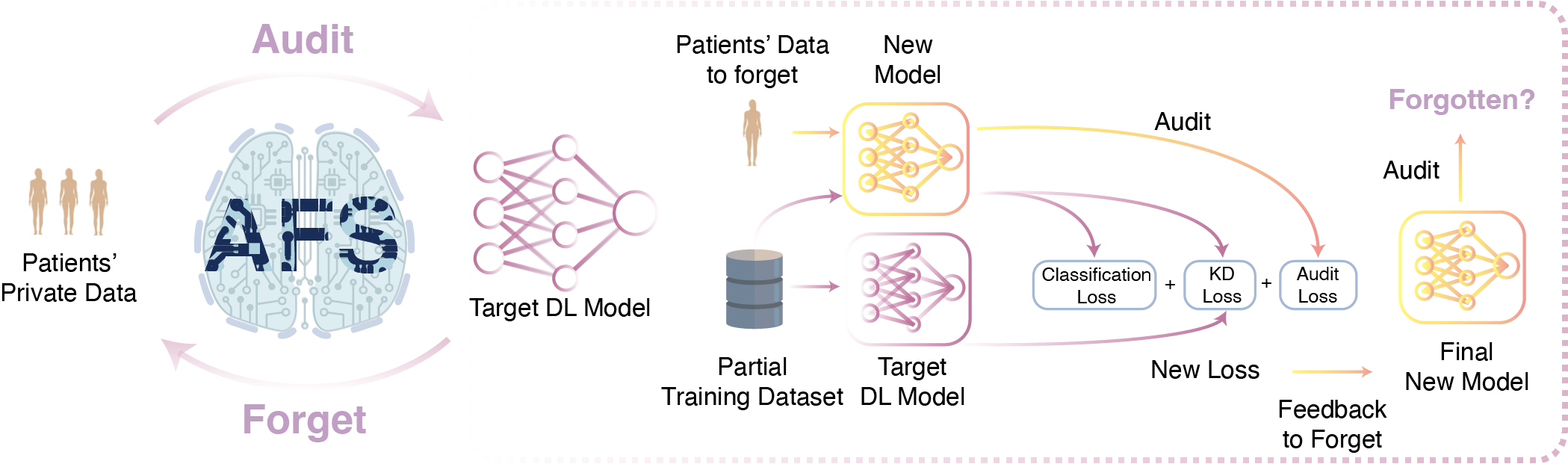}
    \caption{AFS is a unified method to revoke patients’ private data in intelligent healthcare. Given a pre-trained DL model and a query dataset, AFS could audit and provide confidence whether the query dataset has been used to train the target DL model. When a dataset has been used to train the target DL model, AFS could effectively remove the information about the dataset from the target DL model with the guidance of auditing. To achieve that, we proposed a novel method called knowledge purification, which utilizes results from auditing as feedback to forget information.}
    \label{fig1}
\end{figure*}

\begin{figure}[t]
    \centering
    \includegraphics[width=.95\linewidth]{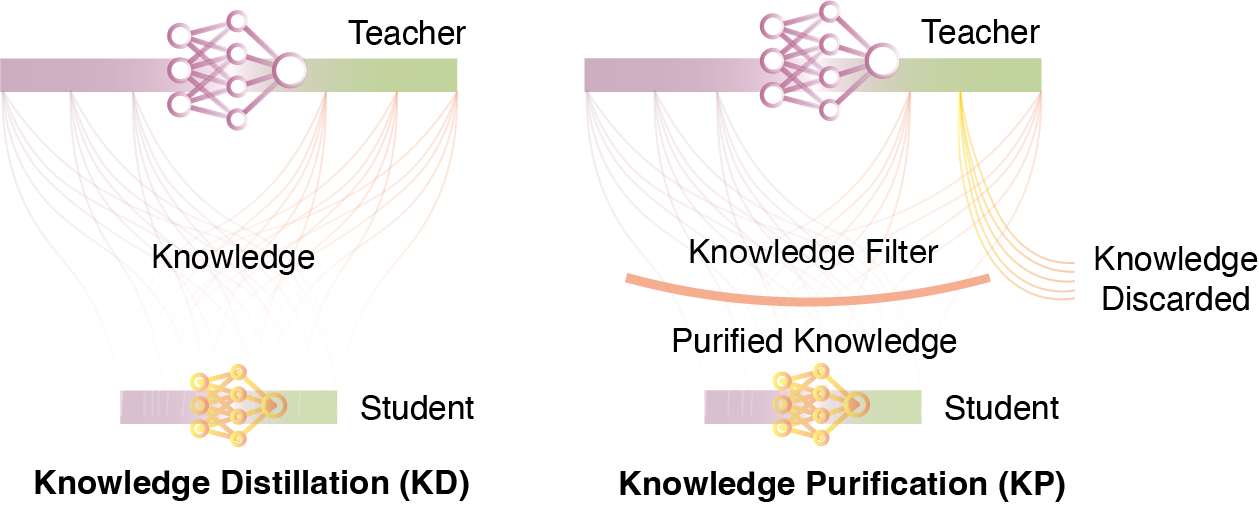}
    \caption{Illustration of knowledge distillation and knowledge purification. Knowledge purification requires the selective transfer of the needed knowledge in the process of knowledge distillation to forget the target information instead of simply transferring all information.}
    \label{fig_kdkp}
\end{figure}

\begin{figure}[t]
    \centering
    \includegraphics[width=.95\linewidth]{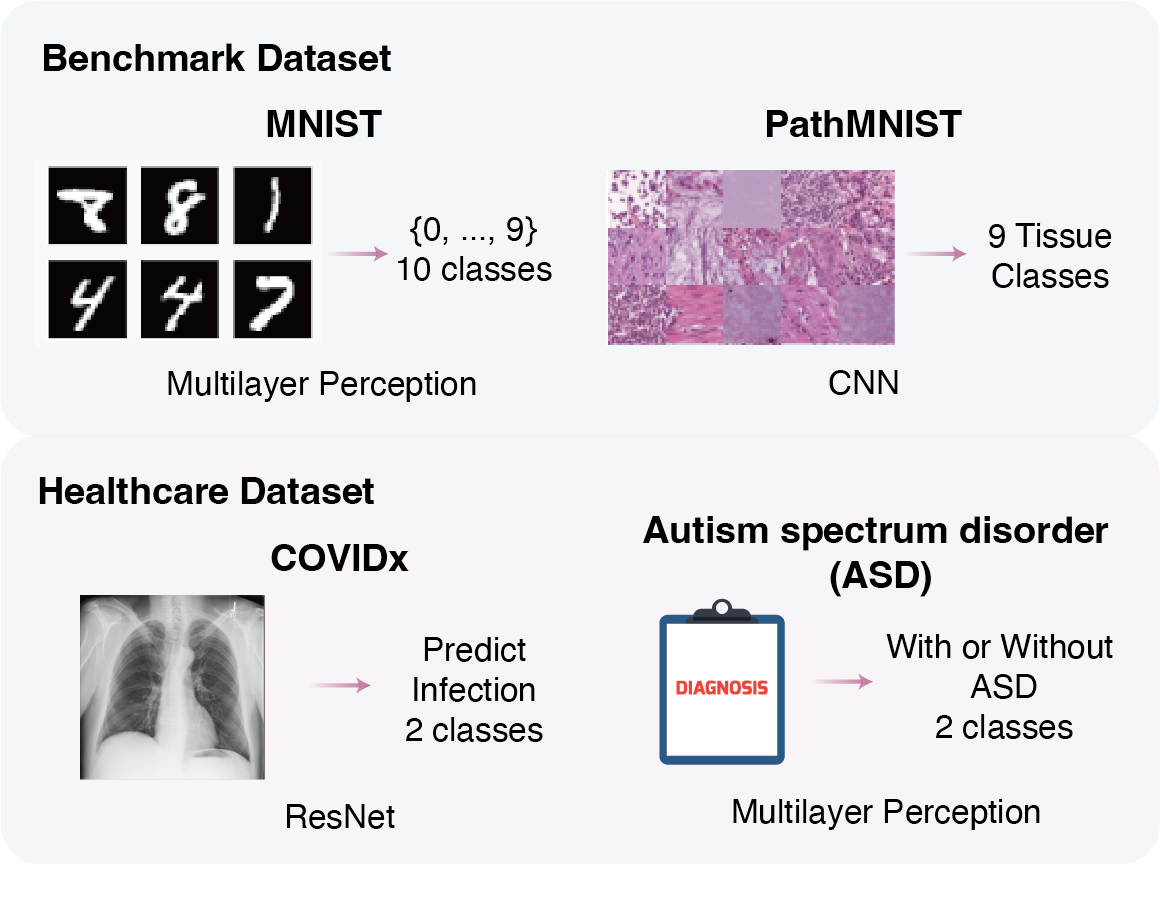}
    \caption{Illustration of four datasets and DL models used to show the versatility of AFS.}
    \label{fig_dataset}
\end{figure}

\section{Methods}

\subsection{The overall framework of AFS}
AFS is a novel and unified method to revoke patients’ private data by using auditing to guide the forgetting process in a negative feedback manner (Figure~\ref{fig1}). 

To audit the membership of the query dataset, AFS takes a pre-trained DL model and the query dataset as inputs, and determines whether the query dataset has been used for training the target DL model. This function was re-implemented based on EMA\cite{huang2021mathsf}, a published MIA-based method to evaluate the membership of a query dataset. Our re-implementation allows quicker and easier usage of auditing (Section \ref{method_audit}). 

To forget the query dataset from a DL model, AFS takes the pre-trained DL model and the query dataset to be forgotten as inputs, in which the query dataset has been used to train the DL model. To effectively forget the information of the query dataset from the pre-trained DL model, an idea is to transfer the information of the remaining dataset except for the query dataset from the pre-trained model to a new model. Therefore, we designed a novel mechanism called knowledge purification (KP) by using auditing to guide the forgetting process to exclude the information of the query dataset while transferring the remaining information by incorporating the auditing loss into the training process (Figure~\ref{fig_kdkp}). With KP integrated, AFS could generate a new model, in which the information of the target dataset should be forgotten under the guidance of auditing (Section \ref{method_forget}). 

To provide an applicable solution, we implemented AFS as open-source software that provides a user-friendly entry point allowing users to use both functions with only one command. To demonstrate the generality of AFS, we applied it to four tasks based on four datasets, including the MNIST dataset, the PathMNIST dataset, the COVIDx dataset and the ASD dataset, which have different data sizes (Figure~\ref{fig_dataset} and Section \ref{method_dataset_preparation}) and various architectures of deep learning networks (Section \ref{method_model}). 

\subsection{Dataset preparation}
\label{method_dataset_preparation}
We used four public datasets that were commonly acknowledged in the machine learning and intelligent healthcare field to demonstrate the versatility of AFS. For the benchmark experiment, we applied AFS on MNIST\cite{lecun1998mnist} and PathMNIST\cite{kather2019predicting} from the MedMNIST\cite{yang2021medmnist} dataset. The MNIST dataset contains 60,000 training images and 10,000 testing images of handwritten digits with size 28×28 and labelled from 0 to 9. PathMNIST contains 100,000 nonoverlapping image patches from hematoxylin \& eosin stained histological images and 7,180 image patches from different clinical centres. In total, 9 types of tissues are involved in the PathMNIST dataset, including adipose, background, debris, lymphocytes, mucus, smooth muscle, normal colon mucosa, cancer-associated stroma, and COAD epithelium. All images in PathMNIST were 224 × 224 (0.5 $\mu$m/px) and were normalized with the Macenko method\cite{macenko2009method}. For the application of AFS in intelligent healthcare, we used the COVIDx\cite{wang2020covid} dataset, which contains 13,975 chest X-ray (CXR) images across 13,870 patient cases, and the Autism spectrum disorder (ASD) dataset for toddlers\cite{thabtah2017autism}, which contains 20 features of 1,054 samples to be utilized for determining influential autistic traits and improving the classification of ASD cases.

For each dataset, we further sampled partial data as the training dataset, the testing dataset, and the calibration dataset as below:

\textbf{MNIST}. We randomly sampled 10,000 images as the training dataset and 10,000 images as the testing dataset. We also randomly sampled 100, 1,000, 2,000, and 5,000 images that are disjoint with the training dataset as four calibration datasets to illustrate the effect of the calibration dataset of varied sizes on auditing and forgetting. 

\textbf{PathMNIST}. We randomly sampled 10,000 images as the training dataset and 5,000 images as the testing dataset. We also randomly sampled 1,000 images that are disjoint with the training dataset as the calibration dataset.

\textbf{COVIDx}. We randomly sampled 5,000 images as the training dataset and 1,000 images as the testing dataset. We also randomly sampled 1,000 images that are disjoint with the training dataset as the calibration dataset.

\textbf{ASD}. We randomly sampled 500 images as the training dataset and 100 images as the testing dataset. We also randomly sampled 100 images that are disjoint with the training dataset as the calibration dataset.

For all four datasets, we randomly sampled partial data from the training dataset with percentage $k$ from \{0.25, 0.5, 0.75\} as the training dataset for knowledge distillation (KD) and AFS. 

In addition, we prepared query datasets with different sizes $N$ from \{1, 10, 100, 500, 1000, 2000\}. A query dataset that completely overlapped with the training dataset is labelled as QO, while the query dataset that is completely disjoint with the training dataset is labelled QNO. To further understand the effect of the purity of the query dataset, we also prepared the query dataset called QM with a $k$ percentage of the query dataset to be overlapped with the training dataset. Finally, for the query dataset designed to be forgotten, we labelled it as QF.

\subsection{Deep learning models and experiment setup}
\label{method_model}
To present the generalizability of AFS towards various DL models, we adopted different architectures for each of the four tasks, including the multilayer perception\cite{gardner1998artificial} (MLP), the convolutional neural network (CNN)\cite{o2015introduction} and ResNet\cite{he2016deep}. There were a large DL model and a small DL model for each task, where the large model refers to the original pre-trained model and the small model is the new model generated by AFS.

For the MNIST dataset, we used MLP with 671,754 parameters as the teacher model and 155,658 parameters as the student model to achieve the 10-class classification task. 

For the PathMNIST dataset, we adopted CNN with 21,285,698 parameters as the teacher model and 11,177,538 parameters as the student network for the 9-class classification task. 

For the COVIDx dataset, we took ResNet34 with 21,285,698 parameters as the teacher model and ResNet18 with 11,177,538 parameters as the student network to achieve the binary classification of healthy people and patients. 

For the ASD dataset, we used the MLP with 3,586 parameters as the teacher model and the MLP with 898 parameters as the student model for the binary classification of autism in toddlers. 

During model training, the number of epochs was fixed to 50, the learning rate was set to 1e–5 and the Adam optimizer was used. A workstation with 252 GB RAM, 112 CPU cores and 2 Nvidia V100 GPUs were adopted for all experiments. The AFS method was developed based on Python3.7, PyTorch1.9.1 and CUDA11.4. A detailed list of dependencies could be found in our code availability. 

\subsection{Audit the membership of query dataset}
\label{method_audit}
EMA\cite{huang2021unlearnable} is designed as a 2-step process. In the first step, the best threshold for each metric is selected to optimize $(TPR(t)+TNR(t))/2$ based on the calibration dataset as shown in Algorithm~\ref{algo_1}. Once the thresholds for all metrics are selected, the membership of each sample in the query dataset will be confirmed as at least one metric is larger than the corresponding threshold. In total, three metrics, including correctness\cite{leino2020stolen}, confidence\cite{yeom2018privacy, song2019privacy}, and negative entropy\cite{ shokri2017membership, salem2018ml}, were adopted in AFS as proposed in the previous work\cite{song2020systematic, huang2021mathsf}. 

Once the membership of all samples in the query dataset is confirmed in the previous step, the query dataset will be further evaluated to determine whether the query dataset has been used to train the target pre-trained DL model. A two-sample statistical test is adopted to evaluate the query dataset based on the sample-wise membership and an all-one vector. The p-value of the two-sample statistical test is used as the output of auditing. Given a user-defined threshold $\alpha$, if $p<\alpha$, then users could conclude that the query dataset was not used for training the target DL model. EMA was re-implemented and integrated into AFS to allow easy and fast auditing.

\begin{algorithm}
    \caption{Infer thresholds}
    \label{algo_1}
    \begin{algorithmic}[1] % The number tells where the line numbering should start
    \Require The calibration dataset $D_{cal}$, the pre-trained DL model $A$ and $n$ different metrics $(m_1, ..., m_n)$ for membership testing.  
    \Procedure{}{} 
        \State Split $D_{cal}$ into training dataset $D_{cal}^{train}$ and test dataset $D_{cal}^{test}$
        \State The calibration model is trained as $f_{D_{cal}}\leftarrow A(D_{cal}^{train})$
        \For{$m_i\in\{m_1, ..., m_n\}$}
            \State Compute metrics for training dataset as $M_{train}\leftarrow\{m_i(f_{D_{cal},s}|s \in D_{cal}^{train})\}$
            \State Compute metrics for test dataset as $M_{test}\leftarrow\{m_i(f_{D_{cal},s}|s \in D_{cal}^{test})\}$
            \State Find $t_i \in argmax_{t\in [M_{train},M_{test}]}(\frac{TPR(t)+TNR(t)}{2})$, where $TPR(t)=\sum_{s\in D_{cal}^{train}}1\{m_i(s)\geq t\}/|D_{cal}^{train}|$ and $TNR(t)=\sum_{s\in D_{cal}^{test}}1\{m_i(s)\geq t\}/|D_{cal}^{test}|$
        \EndFor
    \EndProcedure
    \Return The thresholds $t_1, ..., t_n$ for $n$ metrics
 \end{algorithmic}
\end{algorithm}

\subsection{Audit-guided forgetting of query dataset with AFS}
\label{method_forget}
Forgetting aims to remove the remembered information of the query dataset from the target DL model. Similar to knowledge distillation (KD), a teacher-student paradigm was also adopted in AFS, but with an additional requirement to selectively forget information associated with the data we want to forget. Thus, we designed a novel approach called knowledge purification (KP), meaning purifying the knowledge in the teacher model (the original pre-trained model), discarding the information related to the data that needed to be forgotten and transferring the purified information into the student model (the new model). AFS unified auditing and forgetting into a circular process to effectively enhance the unlearning in a negative feedback manner. 

As shown in Figure~\ref{fig1}, during each epoch of training, the training data will be fed into both the teacher model and the student model, while the data to be forgotten will be audited on the student model. Our main goal is to transfer the knowledge from the teacher model to the student model while forcing the student model to reject the information associated with data to be forgotten. In order to achieve that, we added the audit loss into the total loss, thus allowing the student model to accept partial knowledge from the teacher model and achieve KP as shown in Algorithm~\ref{algo_2}.

\begin{algorithm}
    \caption{AFS}
    \label{algo_2}
    \begin{algorithmic}[1] % The number tells where the line numbering should start
    \Require The calibration dataset $D_{cal}$, the query dataset to forget $D_{forget}$, the sampled training dataset $D_{train}$ for KP, the pre-trained DL model $F$, the new model $f$, and number of epochs $T$.
    \Procedure{}{} 
        \For{epoch $\in \{1, ..., T\}$}
            \State Forward $D_{train}$ with $F$
            \State Forward $D_{train}$ with $f$
            \State Infer threshold with $D_{cal}$ on $f$ and audit $D_{forget}$ on $f$ to get $loss_{audit}$
            \State Calculate $loss_{AFS}=loss_{classification}+loss_{KD}+loss_{audit}$
            \State Update $f$ based on $loss_{AFS}$
        \EndFor
    \EndProcedure
    \Return The new student model $f$ with information about $D_{forget}$ forgotten.
 \end{algorithmic}
\end{algorithm}

\subsection{Evaluation metrices}
Since all four tasks are either multi-classes classification tasks or binary classification tasks, we adopted the accuracy and F1-score as the evaluation metrics as below, 

\begin{equation}
    Accuracy = \frac{TP+TN}{TP+TN+FP+FN}
\end{equation}

\begin{equation}
    F1-score = \frac{2TP}{2TP+FP+FN}
\end{equation}

\noindent where TP represents true positives, TN stands for true negatives, FN represents false negatives and FP stands for false positives.

To evaluate the membership of the query dataset, the p-value of the two-sample statistical test was used as mentioned previously.

%\IEEEpubidadjcol

\begin{figure*}[t]
    \centering
    \includegraphics[width=.95\linewidth]{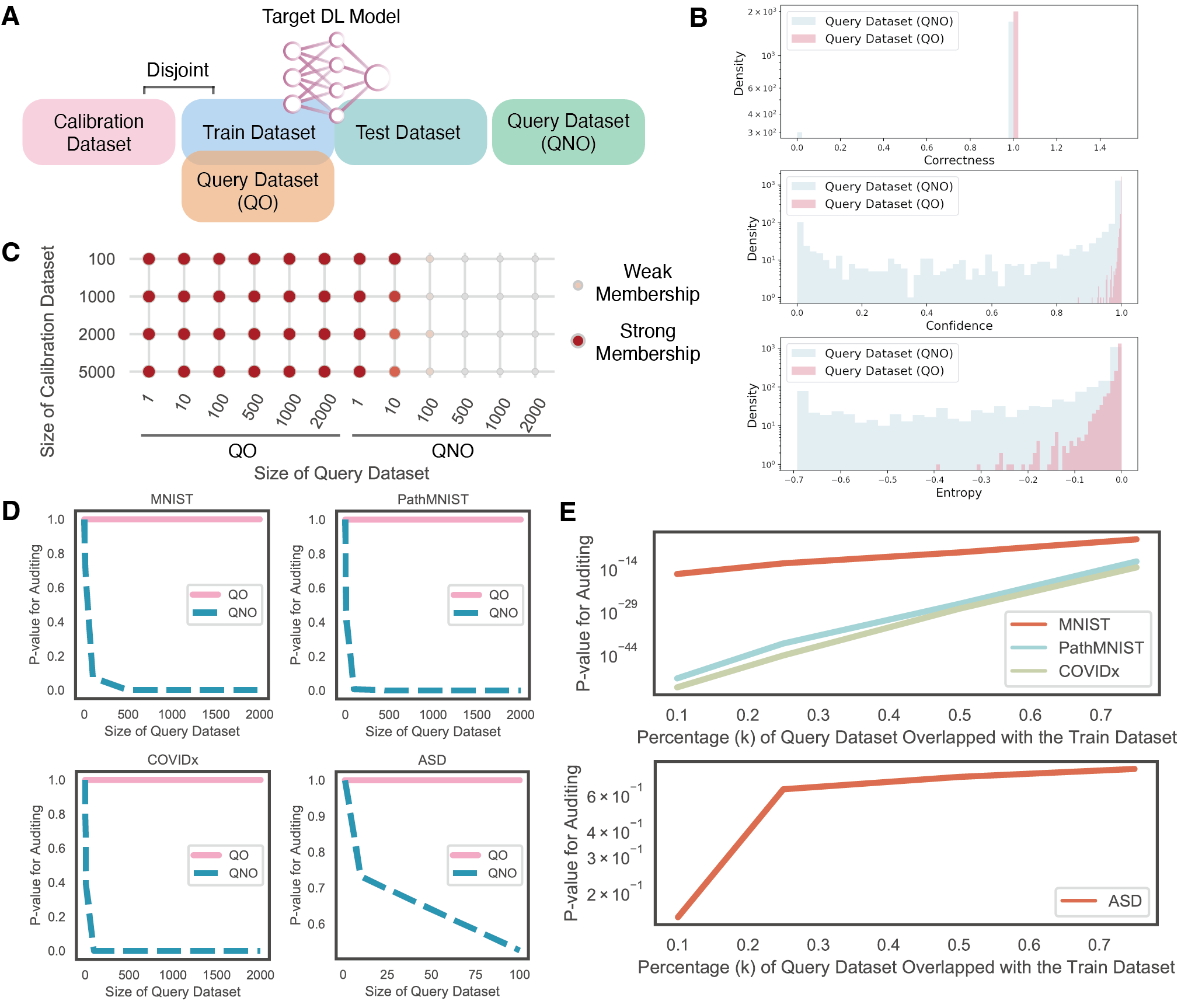}
    \caption{Performance of auditing using AFS on the four datasets. \textbf{A.} Demonstration of the training dataset, the test dataset, the calibration dataset, and the query dataset overlapped with the training dataset (QO) and the query dataset disjointed with the training dataset (QNO). \textbf{B}. Distribution of three metrics for samples in QO and QNO. \textbf{C}. The performance of auditing when varying the size of the calibration dataset and the size of the query dataset. \textbf{D}. The p-value of auditing on QO and QNO of four datasets. \textbf{E}. The p-value of auditing when varying $k$ of the query dataset of four datasets.}
    \label{fig_audit}
\end{figure*}

\section{Results}

\subsection{AFS audits private datasets stably and robustly}
To evaluate the robustness of auditing by AFS, we used it to audit query datasets with different sizes, various purity ($k$ percent of the query dataset was overlapped with the training dataset) and the different sizes of calibration dataset (the size ranged from 100 to 5000) (Method and Figure~\ref{fig_audit}A). For each sample in the query dataset, AFS calculates three metrics for the membership inference, including correctness, confidence and negative entropy (Method). As shown in Figure~\ref{fig_audit}B, all three metrics showed different distributions for QO (query dataset overlapped with the training dataset) and QNO (query dataset disjoint with the training dataset), indicating the dataset-wise divergence of metrics between samples in the training dataset and samples disjoint with the training dataset. Finally, by integrating these three metrics, AFS predicts a p-value to evaluate whether or not a query dataset has been used to train the target DL model. The large p-values indicate the higher probability that the query dataset was used in training. 

When the size of the query dataset and the calibration dataset varied, AFS could still efficiently distinguish QO and QNO (Figure~\ref{fig_audit}C and D). Compared to QO, AFS reported a much smaller p-value for QNO, indicating a weak membership (a small probability that the query dataset has been used to train the target DL model), thus allowing users to judge whether the query dataset was used to train the target DL model. Meanwhile, when the size of the dataset increased from 1 to 2000, AFS discriminated QO and QNO more confidently as there was a more significant divergence of the p-values, which was not affected by the size of the calibration dataset. To further understand the effect of the purity of the query dataset in auditing, we mixed some samples from the training dataset to QNO, thus the new query dataset was labelled as QM (partial data overlapped with the training dataset). The percentage of data overlapped with the training dataset in QM was denoted by $k=\frac{number\ of\ data\ overlapped\ with\ training\ dataset}{size\ of\ QM}$. As shown in Figure~\ref{fig_audit}E, AFS showed a decreasing p-value trend when $k$ decreased, meaning that the query dataset was less likely to be used to train the target DL model. In conclusion, these results indicate the robustness of AFS in determining whether the query data has been used to train the target DL model. 

\begin{table*}[h]
    \centering
        \caption{Comparison of AFS with other methods on auditing QO and QNO from the MNIST dataset with a varied number of samples in the query dataset. The data in the table shows the results of auditing QO and QNO on models trained by different methods. A larger value indicates stronger membership.
        }
        \footnotesize
    \begin{tabular}{l|cccccc|cccccc}
\toprule
\multirow{2}{0pt}{Methods} &  \multicolumn{6}{c|}{QO} & \multicolumn{6}{c}{QNO} \\
 &  1 &  10 &  100 &  500 &  1000 &  2000 &  1 &  10 &  100 &       500 &      1000 &      2000 \\
\midrule
Independent teacher          &    1 &     1 &         1 &         1 &         1 &         1 &     1 &  7.32e-01 &  7.39e-02 &  2.43e-05 &  1.06e-08 &  4.68e-18 \\
Independent student          &    1 &     1 &         1 &  3.43e-01 &  4.22e-01 &  8.61e-02 &     1 &  6.96e-01 &  4.50e-02 &  2.74e-06 &  4.51e-11 &  8.58e-22 \\
\hline
Independent student (k=0.75) &    1 &     1 &         1 &  6.80e-01 &  1.67e-01 &  7.87e-02 &     1 &  5.27e-01 &  2.28e-02 &  7.50e-07 &  4.61e-14 &  3.10e-27 \\
AFS w/o Audit (k=0.75)       &    1 &     1 &         1 &  8.64e-01 &  5.91e-01 &  5.26e-01 &     1 &  5.98e-01 &  1.02e-01 &  6.49e-06 &  4.15e-11 &  1.28e-20 \\
AFS (k=0.75)                 &    1 &     1 &  8.64e-01 &         1 &  8.64e-01 &  4.54e-01 &     1 &  6.96e-01 &  2.94e-02 &  1.45e-06 &  9.06e-13 &  3.79e-24 \\
\hline
Independent student (k=0.5)  &    1 &     1 &  8.64e-01 &  5.59e-01 &  2.39e-01 &  2.49e-02 &     1 &  8.13e-01 &  7.75e-02 &  6.44e-08 &  1.63e-15 &  4.15e-30 \\
AFS w/o Audit (k=0.5)        &    1 &     1 &         1 &  8.64e-01 &  8.64e-01 &  8.64e-01 &     1 &  5.98e-01 &  1.32e-01 &  5.94e-06 &  2.43e-12 &  5.73e-23 \\
AFS (k=0.5)                  &    1 &     1 &         1 &  8.64e-01 &         1 &  7.27e-01 &     1 &  6.68e-01 &  2.88e-02 &  9.25e-07 &  3.70e-13 &  3.10e-27 \\
\hline
Independent student (k=0.25) &    1 &     1 &         1 &  8.64e-01 &  8.64e-01 &  3.17e-01 &     1 &  5.98e-01 &  1.04e-02 &  6.40e-10 &  3.05e-20 &  1.22e-38 \\
AFS w/o Audit (k=0.25)       &    1 &     1 &         1 &         1 &         1 &         1 &     1 &  5.27e-01 &  4.47e-02 &  7.49e-07 &  2.18e-13 &  5.96e-28 \\
AFS (k=0.25)                 &    1 &     1 &         1 &         1 &         1 &         1 &     1 &  6.96e-01 &  8.52e-03 &  3.85e-10 &  1.98e-20 &  1.29e-41 \\
\bottomrule
\end{tabular}
    \label{table_mnist_qo_qno}
\end{table*}

\begin{table*}[h]
    \centering
        \caption{Comparison of AFS with other methods on forgetting QF and model performance with the MNIST dataset. $QF_{100}$ is the small query dataset containing 100 samples and $QF_{1000}$ is the large query dataset containing 1000 samples. We present the p-values of auditing models trained with different methods on $QF_{100}$ and $QF_{1000}$ and the model performance including the accuracy and F1-score.
        }
        \footnotesize
    \begin{tabular}{l|cccccc|cccccc}
\toprule
Methods &    $QF_{100}$ &   $QF_{1000}$ &     Accuracy &      F1-score \\
\midrule
Independent teacher          &         1 &         1 &  0.9622 &  0.9911 \\
Independent student          &         1 &         1 &  0.9504 &  0.9911 \\
\hline
Independent student (k=0.75) &  4.36e-02 &  5.26e-12 &  0.9458 &  0.9880 \\
AFS w/o Audit (k=0.75)       &  3.19e-01 &  1.33e-06 &  0.9582 &  0.9884 \\
AFS (k=0.75)                 &  1.08e-03 &  5.22e-23 &  0.9470 &  0.9889 \\
\hline
Independent student (k=0.5)  &  6.91e-03 &  2.34e-15 &  0.9282 &  0.9848 \\
AFS w/o Audit (k=0.5)        &  1.57e-01 &  7.79e-07 &  0.9526 &  0.9884 \\
AFS (k=0.5)                  &  1.27e-02 &  9.44e-22 &  0.9380 &  0.9866 \\
\hline
Independent student (k=0.25) &  6.91e-03 &  2.90e-19 &  0.9067 &  0.9875 \\
AFS w/o Audit (k=0.25)       &  1.57e-01 &  7.04e-10 &  0.9388 &  0.9875 \\
AFS (k=0.25)                 &  5.79e-04 &  6.84e-33 &  0.9174 &  0.9853 \\
\bottomrule
\end{tabular}
    \label{table_mnist_qf_acc}
\end{table*}

\subsection{AFS forgets the information of query dataset, maintains perfect usability and generates smaller model}
Once the prior knowledge that a dataset has been used to train the target DL model is confirmed with auditing, AFS could be used for forgetting, to remove the information of the dataset from the pre-trained DL model. To comprehensively show the ability of AFS in removing information against the model performance, we compared five methods, including 1) training the teacher model with a complete training dataset (Independent teacher), 2) retraining the student model with a complete training dataset (Independent student), 3) retraining the student model with $k\in\{0.25,0.5,0.75\}$ percentage of the complete training dataset excluding the data to be forgotten (Independent Student with $k\in\{0.25,0.5,0.75\}$), 4) AFS, and 5) training the student model with AFS without the guidance of auditing (AFS w/o Audit), as an ablation study of AFS (Section \ref{method_forget}). Both AFS w/o Audit and AFS were also conducted with varied $k\in\{0.25,0.5,0.75\}$. For both Independent teacher and Independent student methods trained with the complete training dataset, $QF_{100}$ and $QF_{1000}$ were included in the training dataset, while these two query datasets were excluded from the training dataset when $k\in \{0.25, 0.5, 0.75\}$. 

\begin{table*}[h]
    \centering
        \caption{Comparison of AFS with other methods on auditing QO and QNO from the PathMNIST dataset with a varied number of samples in the query dataset. The data in the table shows the results of auditing QO and QNO on models trained by different methods. A larger value indicates stronger membership.
        }
        \footnotesize
    \begin{tabular}{l|cccccc|cccccc}
\toprule
\multirow{2}{0pt}{Methods} &  \multicolumn{6}{c|}{QO} & \multicolumn{6}{c}{QNO} \\
 &  1 &  10 &  100 &  500 &  1000 &  2000 &  1 &  10 &  100 &       500 &      1000 &      2000 \\
\midrule
Independent teacher          &    1 &         1 &         1 &         1 &         1 &         1 &     1 &  4.29e-01 &  6.29e-03 &  1.40e-13 &  4.91e-28 &   1.88e-60 \\
Independent student          &    1 &         1 &         1 &         1 &         1 &         1 &     1 &  5.27e-01 &  5.11e-03 &  1.29e-10 &  3.38e-20 &   7.24e-42 \\
\hline
Independent student (k=0.75) &    1 &         1 &         1 &         1 &         1 &         1 &     1 &  7.32e-01 &  1.89e-02 &  1.55e-08 &  5.74e-16 &   1.92e-30 \\
AFS w/o Audit (k=0.75)       &    1 &         1 &  5.26e-01 &  1.78e-01 &  2.54e-02 &  1.17e-03 &     1 &  2.95e-01 &  2.26e-02 &  9.23e-11 &  1.04e-19 &   8.21e-41 \\
AFS (k=0.75)                 &    1 &         1 &  3.90e-01 &  3.21e-02 &  6.13e-04 &  7.19e-06 &     1 &  4.98e-01 &  2.34e-04 &  4.25e-16 &  3.95e-31 &   4.89e-65 \\
\hline
Independent student (k=0.5)  &    1 &  8.66e-01 &         1 &  8.64e-01 &  8.64e-01 &  5.58e-01 &     1 &  5.62e-01 &  4.02e-03 &  3.01e-10 &  1.91e-21 &   7.85e-40 \\
AFS w/o Audit (k=0.5)        &    1 &         1 &  5.11e-01 &  2.48e-01 &  1.24e-02 &  9.59e-04 &     1 &  6.26e-01 &  1.18e-02 &  2.99e-11 &  4.03e-21 &   2.51e-40 \\
AFS (k=0.5)                  &    1 &         1 &  1.59e-01 &  3.45e-04 &  4.06e-06 &  1.85e-10 &     1 &  2.69e-01 &  9.17e-05 &  2.98e-16 &  3.84e-30 &   1.04e-62 \\
\hline
Independent student (k=0.25) &    1 &         1 &  8.64e-01 &  6.95e-01 &  1.60e-01 &  4.54e-02 &     1 &  1.26e-01 &  1.24e-06 &  2.30e-31 &  2.88e-60 &  4.73e-109 \\
AFS w/o Audit (k=0.25)       &    1 &  8.66e-01 &  2.39e-01 &  3.37e-03 &  1.01e-05 &  9.15e-10 &     1 &  2.42e-01 &  3.64e-05 &  2.55e-24 &  3.01e-43 &   1.37e-90 \\
AFS (k=0.25)                 &    1 &  8.66e-01 &  1.40e-01 &  2.71e-03 &  1.07e-06 &  1.63e-12 &     1 &  4.37e-01 &  7.07e-06 &  2.01e-27 &  4.45e-53 &  1.71e-101 \\
\bottomrule
\end{tabular}
    \label{table_pathmnist_qo_qno}
\end{table*}

\begin{table*}[h]
    \centering
        \caption{Comparison of AFS with other methods on forgetting QF and model performance with the PathMNIST dataset. $QF_{100}$ is the small query dataset containing 100 samples and $QF_{1000}$ is the large query dataset containing 1000 samples. We present the p-values of auditing models trained with different methods on $QF_{100}$ and $QF_{1000}$ and the model performance including the accuracy and F1-score.
        }
        \footnotesize
    \begin{tabular}{l|cccccc|cccccc}
\toprule
Methods &    $QF_{100}$ &   $QF_{1000}$ &     Accuracy &      F1-score \\
\midrule
Independent teacher          &         1 &         1 &  0.8538 &  0.9885 \\
Independent student          &         1 &         1 &  0.8446 &  0.9836 \\
\hline
Independent student (k=0.75) &  2.35e-02 &  4.08e-15 &  0.8396 &  0.9555 \\
AFS w/o Audit (k=0.75)       &  1.08e-03 &  1.67e-22 &  0.8682 &  0.9777 \\
AFS (k=0.75)                 &  2.25e-05 &  2.05e-41 &  0.8560 &  0.9605 \\
\hline
Independent student (k=0.5)  &  6.91e-03 &  2.99e-21 &  0.7934 &  0.9533 \\
AFS w/o Audit (k=0.5)        &  3.74e-03 &  4.93e-18 &  0.8494 &  0.9697 \\
AFS (k=0.5)                  &  2.87e-06 &  4.75e-35 &  0.8242 &  0.9575 \\
\hline
Independent student (k=0.25) &  1.58e-07 &  9.05e-57 &  0.7582 &  0.9287 \\
AFS w/o Audit (k=0.25)       &  3.32e-07 &  2.05e-41 &  0.7842 &  0.9406 \\
AFS (k=0.25)                 &  3.32e-07 &  1.84e-56 &  0.7810 &  0.9385 \\
\bottomrule
\end{tabular}
    \label{table_pathmnist_qf_acc}
\end{table*}

Taking the MNIST dataset as an example, for models trained with each method, except for auditing on QO and QNO, we further audited the membership of two datasets designed to be forgotten (a small query dataset $QF_{100}$ and a large query dataset $QF_{1000}$) to assess the ability of different methods in forgetting the query dataset. As shown in Table~\ref{table_mnist_qo_qno}, regardless of the model trained based on which method, AFS could effectively distinguish between QO and QNO, and the divergence in auditing two query datasets was enlarged as the size of the query dataset increased.

As shown in Table~\ref{table_mnist_qf_acc}, AFS perfectly predicted the membership of $QF_{100}$ and $QF_{1000}$ on both models from Independent teacher and Independent student methods as both query datasets were included in the training dataset. Since both query datasets were disjoint with the partial training dataset when $k\in\{0.25,0.5,0.75\}$, thus auditing on the model trained with Independent student with $k\in\{0.25,0.5,0.75\}$ weakly denied the membership of $QF_{100}$ ($P_{QF100,   k=0.75}=4.36E–2$, $P_{QF100,   k=0.5}=6.91E–3$, $P_{QF100,   k=0.25}=6.91E–3)$ and $QF_{1000}$ ($P_{QF1000,   k=0.75}=5.26E–12$, $P_{QF1000,   k=0.5}=2.34E–15$, $P_{QF1000,   k=0.25}=2.90E–19$). However, since only the partial training dataset was used when $k\in\{0.25,0.5,0.75\}$, the retrained models with Independent student only learnt the information of the partial training dataset and lost the information from the remaining data in the complete training dataset, thus resulting in the significant drop of model performance compared to either the Independent student or the Independent teacher trained with the complete training dataset.

To rescue the information lost due to the usage of partial training samples and further increase the model performance, AFS could use only a partial training dataset ($k\in\{0.25,0.5,0.75\}$) to transfer the knowledge from the Independent teacher pre-trained with the complete training dataset. As shown in Table~\ref{table_mnist_qf_acc}, the model trained with AFS provided higher accuracy and F1-score compared to the Independent student trained with partial training dataset ($k\in\{0.25,0.5,0.75\}$) and together with a better forgetting performance (much smaller auditing score on $QF_{100}$ and $QF_{1000}$), as AFS used auditing as feedback for forgetting and could forget not only the query samples but also other samples with similar features. 

\begin{figure*}[t]
    \centering
    \includegraphics[width=.95\linewidth]{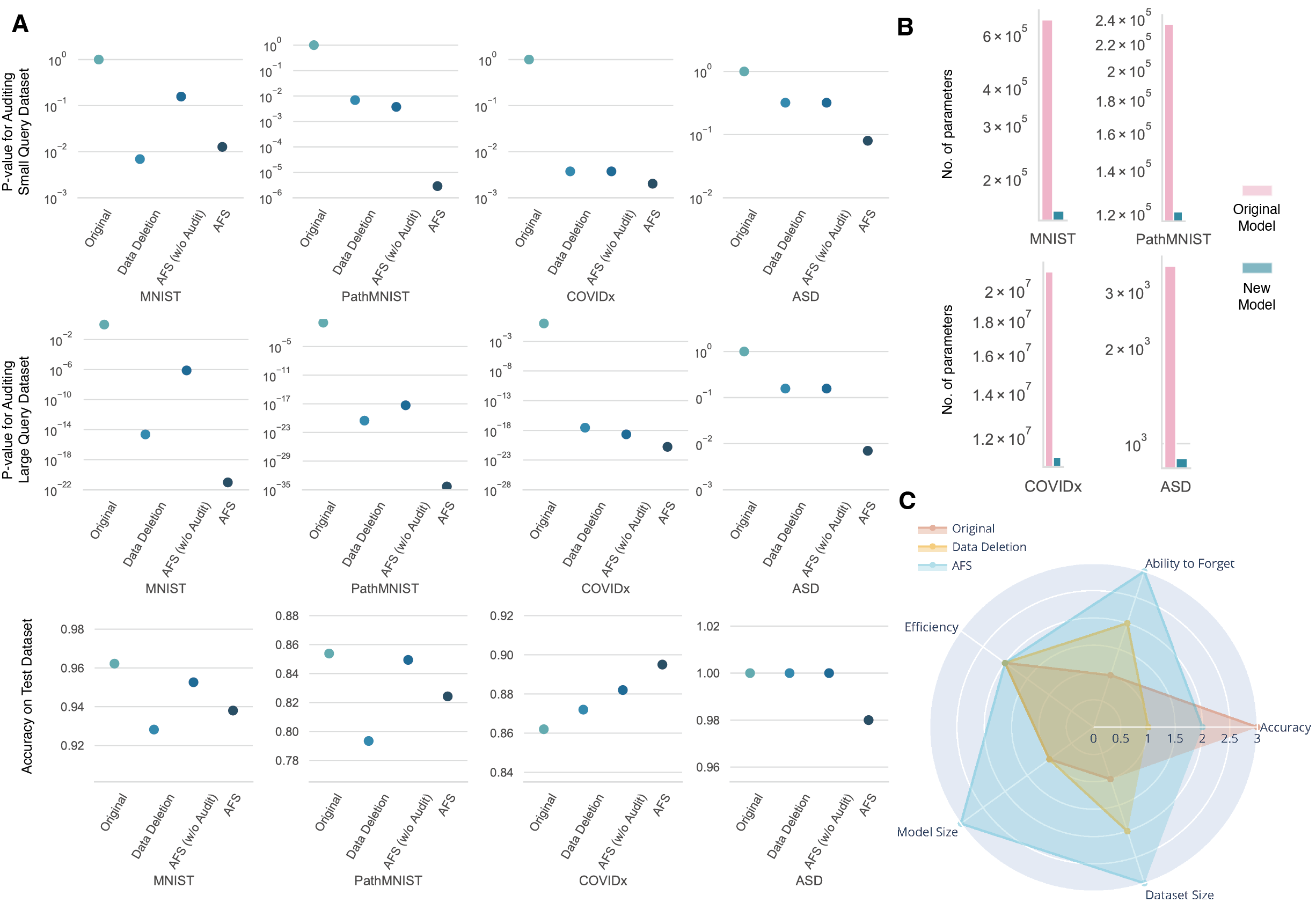}
    \caption{Performance of forgetting using AFS on four datasets. \textbf{A.} The p-value of auditing on a small query dataset and a large query dataset (QF) and the accuracy of models trained with different methods, including Original (Independent teacher trained with the complete training dataset), Data Deletion (the Independent student model trained with partial training dataset and $k=0.5$), AFS (w/o Audit) and AFS. \textbf{B.} The number of parameters for the original large model and the new small model generated by AFS. \textbf{C.} The qualitative evaluation of three methods, including Original (Independent teacher trained with the complete training dataset), Data Deletion (the Independent student model trained with partial training dataset and $k=0.5$), and AFS on five dimensions (Ability to forget, accuracy, size of dataset needed for training, size of the generated model and the efficiency of training). A larger value means a stronger ability to forget, higher model accuracy, a smaller size of dataset needed for training, a smaller size of the generated model, and better efficiency of training.}
    \label{fig_performance_4_datasets}
\end{figure*}

We also applied AFS on the 9-classes classification of hematoxylin \& eosin-stained histological images from the PathMNIST dataset with CNN. As shown in Table~\ref{table_pathmnist_qo_qno}, AFS could still distinguish QO and QNO from the PathMNIST dataset. The divergence of auditing between QO and QNO was more significant than that on the MNIST dataset. With the requirement to forget both query datasets ($QF_{100}$ and $QF_{1000}$), the model trained with AFS outperformed on forgetting information ($P_{QF100,   k=0.75}=2.25E–5$, $P_{QF100,   k=0.5}=2.87E–6$, $P_{QF100,   k=0.25}=3.32E–7)$, $P_{QF1000,   k=0.75}=2.05E–41$, $P_{QF1000,   k=0.5}=4.75E–35$, $P_{QF1000,   k=0.25}=1.84E–56$) while learnt more information from the Independent teacher model trained with a complete training dataset.

In summary, AFS could effectively forget the information of the query dataset from the target DL model. Since KP was integrated into AFS, it could generate a smaller DL model, which masters knowledge from the larger teacher model by using only a partial training dataset ($k=0.5$ could achieve a good balance between forgetting and model performance), without the need to retrain the larger model with the complete training dataset. Compared to retraining the student model, the model trained with AFS showed even better performance in forgetting the information while maintaining better model performance (accuracy and F1-score) as it learnt the knowledge from the model trained with the complete training dataset. As shown by the ablation study in Tables~\ref{table_mnist_qf_acc} and ~\ref{table_pathmnist_qf_acc}, compared to AFS w/o Audit, the audit-guided AFS could forget the information more significantly but with an acceptable cost in decreasing the model performance (accuracy and f1-score).

\subsection{Apply AFS to forget medical images}
To show the versatility of AFS, we applied it to the classification of pneumonia and normal with chest X-ray images from the COVIDx dataset with ResNet, which is a classic task in medical image analysis. As shown in Figure~\ref{fig_performance_4_datasets}A, on both query datasets ($QF_{100}$ and $QF_{1000}$), AFS could effectively forget the information of the query dataset, while generating the new model with much less number of parameters as shown in Figure~\ref{fig_performance_4_datasets}B. Surprisingly, the model generated by AFS showed even better accuracy than the Independent teacher trained with the complete dataset and the Independent student trained with the partial training dataset. This result not only indicated that AFS could effectively transfer the knowledge from the teacher model to the student model but also suggested that the student model with simpler architecture could even perform better than the teacher model with KP in AFS due to the reduction of model parameters and purification of knowledge in some real-world cases.

\subsection{Apply AFS to forget electrical health records}

To further prove the generalizability of AFS in both the auditing and forgetting, we applied AFS to predicting early autism spectrum disorder (ASD) traits of toddlers, which contains sensitive information about patients, such as the age, gender and the family gene trait. That information was stored as electrical health records (EHR). As shown in Figure~\ref{fig_performance_4_datasets}A, similar to previous results on other datasets, AFS effectively removed the information of both query datasets from the pre-trained DL model. Since the size of the ASD dataset was quite small, we adopted two smaller query datasets ($QF_{50}$ and $QF_{100}$) to be forgotten. Compared to the models trained with other methods, the model trained with AFS successfully forgot the information of both $QF_{50}$ ($P_{QF_{50},   k=0.75}=0.08$, $P_{QF_{50},   k=0.55}=0.08$, $P_{QF_{50},   k=0.25}=0.156$) and $QF_{100}$ ($P_{QF_{100},   k=0.75}=0.004$, $P_{QF_{100},   k=0.55}=0.007$, $P_{QF_{100},   k=0.25}=0.007$) without affecting the model utility significantly ($Acc_{AFS,k=0.75}=0.98$, $Acc_{AFS,k=0.5}=0.98$, $Acc_{AFS,k=0.25}=0.98$).

\begin{table}[h]
    \centering
        \caption{Time for inferring 100 samples with the original model and the model generated by AFS.
        }
        \footnotesize
    \begin{tabular}{c|c|c}
\toprule
Dataset & Original model & New model generated by AFS \\
\midrule
MNIST          &    439 µs ± 1.54 µs &  284 µs ± 447 ns \\
\hline
PathMNIST & 5.13 ms ± 22 µs & 4.99 ms ± 14.1 µs\\
\hline
COVIDx  & 1.27 s ± 500 ms & 661 ms ± 9.98 ms \\
\hline
Autism & 126 µs ± 177 ns & 87.3 µs ± 120 ns\\
\bottomrule
\end{tabular}
    \label{table_infer_time}
\end{table}

\begin{table}[h]
    \centering
        \caption{GPU memory (MB) for inferring 100 samples with the original model and the model generated by AFS.
        }
        \footnotesize
    \begin{tabular}{c|c|c}
\toprule
Dataset & Original model & New model generated by AFS \\
\midrule
MNIST & 258 & 61  \\
\hline
PathMNIST & 173 & 129 \\
\hline
COVIDx  & 17,805 & 10,600 \\
\hline
Autism & 2 & 1 \\
\bottomrule
\end{tabular}
    \label{table_infer_GPU}
\end{table}

\section{Discussion}
To our knowledge, AFS is the first unified method of auditing and forgetting that could effectively forget the information of the target query dataset from the pre-trained DL model with the guidance of auditing. We designed AFS as a model-agnostic and open-source method that is applicable to different models. As shown in Figure~\ref{fig_performance_4_datasets}C, AFS could generate a smaller model, which requires much less time and GPU memory during the inference (Tables~\ref{table_infer_time} and \ref{table_infer_GPU}), by training with a partial training dataset ($\sim$50\%) with our novel KP approach. Moreover, AFS could forget the information of the query dataset at the expense of an acceptable reduction in the model performance.

Our experiments on four datasets showed that AFS was generalized for datasets of different sizes and forms, including medical images and EHR. Since deep learning models with different architectures were applied to four tasks, we further demonstrated the broad applicability of AFS to common deep learning models. In addition, our tasks include both binary classification and multiclassification tasks, which also suggested that AFS was applicable for tasks with multiple labels.

With current laws that guarantee people the right to revoke their own data, AFS could help institutions and companies to efficiently iterate their models to forget individual information at the model level. However, there are still some shortcomings in the application of the current version of AFS in the production environment, which could be the main potential direction of research in the future. Firstly, the models and data we tested in this study were still not large enough compared to the data in the real production environment. Therefore, it is unknown whether scaling AFS to larger models and more data will cause new problems. Secondly, there are different approaches to audit, and thus we could add more metrics of auditing to AFS to guide the forgetting process in the future version. Finally, due to the limitation of auditing, it is still difficult to perform individual-level forgetting, as we need to compare the difference in statistical distribution based on a fraction of data points, which could be the major possible improvement for the future version of AFS. Despite these limitations, we believe that AFS will make a valuable contribution towards better protection of people's privacy and the right to revoke the data with the rapid development of intelligent healthcare.

\section{Data availability}
All four datasets used in this work are publicly available. The MNIST dataset could be found at \url{http://yann.lecun.com/exdb/mnist/}. The PathMNIST dataset is available at \url{https://medmnist.com/}. The COVIDx dataset is stored at \url{https://www.kaggle.com/datasets/andyczhao/covidx-cxr2?select=competition_test}. The ASD dataset can be accessed at \url{https://www.kaggle.com/datasets/fabdelja/autism-screening-for-toddlers}.

\section{Code availability}
The AFS software is publicly available at \url{https://github.com/JoshuaChou2018/AFS}

\section{Credit author statement}
Conceptualization: J.Z. and X.G. Design: J.Z. and X.G. Data analysis and interpretation: J.Z., H.L. and W.H. Code implementation: J.Z. and H.L. Application: J.Z, H.L, X.L, B.Z. Code improvement: Z.L and L.Z. Drafting of the manuscript: J.Z and H.L. Critical revision of the manuscript for important intellectual content: J.Z., X.L and B.Z. Supervision: J.Z. and X.G. Funding acquisition: X.G.

\section{Acknowledgements}
Juexiao Zhou, Haoyang Li, Xingyu Liao, Bin Zhang, Wenjia He, Zhongxiao Li, Longxi Zhou and Xin Gao were supported in part by grants from the Office of Research Administration (ORA) at King Abdullah University of Science and Technology (KAUST) under award number FCC/1/1976-44-01, FCC/1/1976-45-01, REI/1/5202-01-01, REI/1/5234-01-01, REI/1/4940-01-01, RGC/3/4816-01-01, and REI/1/0018-01-01.

Xingyu Liao was also supported in part by grants from the National Natural Science Foundation of China under grant number No.62002388.

\section{Competing Interests}
The authors have declared no competing interests.

{
\bibliographystyle{IEEEtran}
\bibliography{reg}

\begin{thebibliography}{10}
\providecommand{\url}[1]{#1}
\csname url@samestyle\endcsname
\providecommand{\newblock}{\relax}
\providecommand{\bibinfo}[2]{#2}
\providecommand{\BIBentrySTDinterwordspacing}{\spaceskip=0pt\relax}
\providecommand{\BIBentryALTinterwordstretchfactor}{4}
\providecommand{\BIBentryALTinterwordspacing}{\spaceskip=\fontdimen2\font plus
\BIBentryALTinterwordstretchfactor\fontdimen3\font minus
  \fontdimen4\font\relax}
\providecommand{\BIBforeignlanguage}[2]{{%
\expandafter\ifx\csname l@#1\endcsname\relax
\typeout{** WARNING: IEEEtran.bst: No hyphenation pattern has been}%
\typeout{** loaded for the language `#1'. Using the pattern for}%
\typeout{** the default language instead.}%
\else
\language=\csname l@#1\endcsname
\fi
#2}}
\providecommand{\BIBdecl}{\relax}
\BIBdecl

\bibitem{voigt2017eu}
P.~Voigt and A.~Von~dem Bussche, ``The eu general data protection regulation
  (gdpr),'' \emph{A Practical Guide, 1st Ed., Cham: Springer International
  Publishing}, vol.~10, no. 3152676, pp. 10--5555, 2017.

\bibitem{act1996health}
A.~Act, ``Health insurance portability and accountability act of 1996,''
  \emph{Public law}, vol. 104, p. 191, 1996.

\bibitem{pardau2018california}
S.~L. Pardau, ``The california consumer privacy act: Towards a european-style
  privacy regime in the united states,'' \emph{J. Tech. L. \& Pol'y}, vol.~23,
  p.~68, 2018.

\bibitem{wang2009learning}
R.~Wang, Y.~F. Li, X.~Wang, H.~Tang, and X.~Zhou, ``Learning your identity and
  disease from research papers: information leaks in genome wide association
  study,'' in \emph{Proceedings of the 16th ACM conference on Computer and
  communications security}, 2009, pp. 534--544.

\bibitem{fredrikson2014privacy}
M.~Fredrikson, E.~Lantz, S.~Jha, S.~Lin, D.~Page, and T.~Ristenpart, ``Privacy
  in pharmacogenetics: An $\{$End-to-End$\}$ case study of personalized
  warfarin dosing,'' in \emph{23rd USENIX Security Symposium (USENIX Security
  14)}, 2014, pp. 17--32.

\bibitem{cao2015towards}
Y.~Cao and J.~Yang, ``Towards making systems forget with machine unlearning,''
  in \emph{2015 IEEE Symposium on Security and Privacy}.\hskip 1em plus 0.5em
  minus 0.4em\relax IEEE, 2015, pp. 463--480.

\bibitem{fredrikson2015model}
M.~Fredrikson, S.~Jha, and T.~Ristenpart, ``Model inversion attacks that
  exploit confidence information and basic countermeasures,'' in
  \emph{Proceedings of the 22nd ACM SIGSAC conference on computer and
  communications security}, 2015, pp. 1322--1333.

\bibitem{song2017machine}
C.~Song, T.~Ristenpart, and V.~Shmatikov, ``Machine learning models that
  remember too much,'' in \emph{Proceedings of the 2017 ACM SIGSAC Conference
  on computer and communications security}, 2017, pp. 587--601.

\bibitem{ganju2018property}
K.~Ganju, Q.~Wang, W.~Yang, C.~A. Gunter, and N.~Borisov, ``Property inference
  attacks on fully connected neural networks using permutation invariant
  representations,'' in \emph{Proceedings of the 2018 ACM SIGSAC conference on
  computer and communications security}, 2018, pp. 619--633.

\bibitem{carlini2019secret}
N.~Carlini, C.~Liu, {\'U}.~Erlingsson, J.~Kos, and D.~Song, ``The secret
  sharer: Evaluating and testing unintended memorization in neural networks,''
  in \emph{28th USENIX Security Symposium (USENIX Security 19)}, 2019, pp.
  267--284.

\bibitem{zhou2022ppml}
J.~Zhou, S.~Chen, Y.~Wu, H.~Li, B.~Zhang, L.~Zhou, Y.~Hu, Z.~Xiang, Z.~Li,
  N.~Chen \emph{et~al.}, ``Ppml-omics: a privacy-preserving federated machine
  learning system protects patients’ privacy from omic data,''
  \emph{bioRxiv}, 2022.

\bibitem{mckinney2020international}
S.~M. McKinney, M.~Sieniek, V.~Godbole, J.~Godwin, N.~Antropova, H.~Ashrafian,
  T.~Back, M.~Chesus, G.~S. Corrado, A.~Darzi \emph{et~al.}, ``International
  evaluation of an ai system for breast cancer screening,'' \emph{Nature}, vol.
  577, no. 7788, pp. 89--94, 2020.

\bibitem{ardila2019end}
D.~Ardila, A.~P. Kiraly, S.~Bharadwaj, B.~Choi, J.~J. Reicher, L.~Peng, D.~Tse,
  M.~Etemadi, W.~Ye, G.~Corrado \emph{et~al.}, ``End-to-end lung cancer
  screening with three-dimensional deep learning on low-dose chest computed
  tomography,'' \emph{Nature medicine}, vol.~25, no.~6, pp. 954--961, 2019.

\bibitem{poplin2018prediction}
R.~Poplin, A.~V. Varadarajan, K.~Blumer, Y.~Liu, M.~V. McConnell, G.~S.
  Corrado, L.~Peng, and D.~R. Webster, ``Prediction of cardiovascular risk
  factors from retinal fundus photographs via deep learning,'' \emph{Nature
  Biomedical Engineering}, vol.~2, no.~3, pp. 158--164, 2018.

\bibitem{zhou2020rapid}
L.~Zhou, Z.~Li, J.~Zhou, H.~Li, Y.~Chen, Y.~Huang, D.~Xie, L.~Zhao, M.~Fan,
  S.~Hashmi \emph{et~al.}, ``A rapid, accurate and machine-agnostic
  segmentation and quantification method for ct-based covid-19 diagnosis,''
  \emph{IEEE transactions on medical imaging}, vol.~39, no.~8, pp. 2638--2652,
  2020.

\bibitem{zhou2022interpretable}
L.~Zhou, X.~Meng, Y.~Huang, K.~Kang, J.~Zhou, Y.~Chu, H.~Li, D.~Xie, J.~Zhang,
  W.~Yang \emph{et~al.}, ``An interpretable deep learning workflow for
  discovering subvisual abnormalities in ct scans of covid-19 inpatients and
  survivors,'' \emph{Nature Machine Intelligence}, vol.~4, no.~5, pp. 494--503,
  2022.

\bibitem{bartoletti2019ai}
I.~Bartoletti, ``Ai in healthcare: Ethical and privacy challenges,'' in
  \emph{Conference on Artificial Intelligence in Medicine in Europe}.\hskip 1em
  plus 0.5em minus 0.4em\relax Springer, 2019, pp. 7--10.

\bibitem{bourtoule2021machine}
L.~Bourtoule, V.~Chandrasekaran, C.~A. Choquette-Choo, H.~Jia, A.~Travers,
  B.~Zhang, D.~Lie, and N.~Papernot, ``Machine unlearning,'' in \emph{2021 IEEE
  Symposium on Security and Privacy (SP)}.\hskip 1em plus 0.5em minus
  0.4em\relax IEEE, 2021, pp. 141--159.

\bibitem{nguyen2020variational}
Q.~P. Nguyen, B.~K.~H. Low, and P.~Jaillet, ``Variational bayesian
  unlearning,'' \emph{Advances in Neural Information Processing Systems},
  vol.~33, pp. 16\,025--16\,036, 2020.

\bibitem{nguyen2022survey}
T.~T. Nguyen, T.~T. Huynh, P.~L. Nguyen, A.~W.-C. Liew, H.~Yin, and Q.~V.~H.
  Nguyen, ``A survey of machine unlearning,'' \emph{arXiv preprint
  arXiv:2209.02299}, 2022.

\bibitem{gupta2021adaptive}
V.~Gupta, C.~Jung, S.~Neel, A.~Roth, S.~Sharifi-Malvajerdi, and C.~Waites,
  ``Adaptive machine unlearning,'' \emph{Advances in Neural Information
  Processing Systems}, vol.~34, pp. 16\,319--16\,330, 2021.

\bibitem{thudi2022unrolling}
A.~Thudi, G.~Deza, V.~Chandrasekaran, and N.~Papernot, ``Unrolling sgd:
  Understanding factors influencing machine unlearning,'' in \emph{2022 IEEE
  7th European Symposium on Security and Privacy (EuroS\&P)}.\hskip 1em plus
  0.5em minus 0.4em\relax IEEE, 2022, pp. 303--319.

\bibitem{guo2019certified}
C.~Guo, T.~Goldstein, A.~Hannun, and L.~Van Der~Maaten, ``Certified data
  removal from machine learning models,'' \emph{arXiv preprint
  arXiv:1911.03030}, 2019.

\bibitem{golatkar2020eternal}
A.~Golatkar, A.~Achille, and S.~Soatto, ``Eternal sunshine of the spotless net:
  Selective forgetting in deep networks,'' in \emph{Proceedings of the IEEE/CVF
  Conference on Computer Vision and Pattern Recognition}, 2020, pp. 9304--9312.

\bibitem{neel2021descent}
S.~Neel, A.~Roth, and S.~Sharifi-Malvajerdi, ``Descent-to-delete:
  Gradient-based methods for machine unlearning,'' in \emph{Algorithmic
  Learning Theory}.\hskip 1em plus 0.5em minus 0.4em\relax PMLR, 2021, pp.
  931--962.

\bibitem{ginart2019making}
A.~Ginart, M.~Guan, G.~Valiant, and J.~Y. Zou, ``Making ai forget you: Data
  deletion in machine learning,'' \emph{Advances in neural information
  processing systems}, vol.~32, 2019.

\bibitem{chundawat2022can}
V.~S. Chundawat, A.~K. Tarun, M.~Mandal, and M.~Kankanhalli, ``Can bad teaching
  induce forgetting? unlearning in deep networks using an incompetent
  teacher,'' \emph{arXiv preprint arXiv:2205.08096}, 2022.

\bibitem{kim2022efficient}
J.~Kim and S.~S. Woo, ``Efficient two-stage model retraining for machine
  unlearning,'' in \emph{Proceedings of the IEEE/CVF Conference on Computer
  Vision and Pattern Recognition}, 2022, pp. 4361--4369.

\bibitem{nguyen2022markov}
Q.~P. Nguyen, R.~Oikawa, D.~M. Divakaran, M.~C. Chan, and B.~K.~H. Low,
  ``Markov chain monte carlo-based machine unlearning: Unlearning what needs to
  be forgotten,'' \emph{arXiv preprint arXiv:2202.13585}, 2022.

\bibitem{baumhauer2022machine}
T.~Baumhauer, P.~Sch{\"o}ttle, and M.~Zeppelzauer, ``Machine unlearning: Linear
  filtration for logit-based classifiers,'' \emph{Machine Learning}, vol. 111,
  no.~9, pp. 3203--3226, 2022.

\bibitem{izzo2021approximate}
Z.~Izzo, M.~A. Smart, K.~Chaudhuri, and J.~Zou, ``Approximate data deletion
  from machine learning models,'' in \emph{International Conference on
  Artificial Intelligence and Statistics}.\hskip 1em plus 0.5em minus
  0.4em\relax PMLR, 2021, pp. 2008--2016.

\bibitem{schelter2021hedgecut}
S.~Schelter, S.~Grafberger, and T.~Dunning, ``Hedgecut: Maintaining randomised
  trees for low-latency machine unlearning,'' in \emph{Proceedings of the 2021
  International Conference on Management of Data}, 2021, pp. 1545--1557.

\bibitem{shan2020protecting}
S.~Shan, E.~Wenger, J.~Zhang, H.~Li, H.~Zheng, and B.~Zhao, ``Protecting
  personal privacy against una uthorized deep learning models,'' in
  \emph{Proceedings of USENIX Security Symposium}, 2020, pp. 1--16.

\bibitem{tarun2021fast}
A.~K. Tarun, V.~S. Chundawat, M.~Mandal, and M.~Kankanhalli, ``Fast yet
  effective machine unlearning,'' \emph{arXiv preprint arXiv:2111.08947}, 2021.

\bibitem{huang2021unlearnable}
H.~Huang, X.~Ma, S.~M. Erfani, J.~Bailey, and Y.~Wang, ``Unlearnable examples:
  Making personal data unexploitable,'' \emph{arXiv preprint arXiv:2101.04898},
  2021.

\bibitem{peste2021ssse}
A.~Peste, D.~Alistarh, and C.~H. Lampert, ``Ssse: Efficiently erasing samples
  from trained machine learning models,'' \emph{arXiv preprint
  arXiv:2107.03860}, 2021.

\bibitem{liu2020have}
X.~Liu and S.~A. Tsaftaris, ``Have you forgotten? a method to assess if machine
  learning models have forgotten data,'' in \emph{International Conference on
  Medical Image Computing and Computer-Assisted Intervention}.\hskip 1em plus
  0.5em minus 0.4em\relax Springer, 2020, pp. 95--105.

\bibitem{huang2021mathsf}
Y.~Huang, X.~Li, and K.~Li, ``Ema: Auditing data removal from trained models,''
  in \emph{International Conference on Medical Image Computing and
  Computer-Assisted Intervention}.\hskip 1em plus 0.5em minus 0.4em\relax
  Springer, 2021, pp. 793--803.

\bibitem{hullermeier2021aleatoric}
E.~H{\"u}llermeier and W.~Waegeman, ``Aleatoric and epistemic uncertainty in
  machine learning: An introduction to concepts and methods,'' \emph{Machine
  Learning}, vol. 110, no.~3, pp. 457--506, 2021.

\bibitem{hinton2015distilling}
G.~Hinton, O.~Vinyals, J.~Dean \emph{et~al.}, ``Distilling the knowledge in a
  neural network,'' \emph{arXiv preprint arXiv:1503.02531}, vol.~2, no.~7,
  2015.

\bibitem{lecun1998mnist}
Y.~LeCun, ``The mnist database of handwritten digits,'' \emph{http://yann.
  lecun. com/exdb/mnist/}, 1998.

\bibitem{kather2019predicting}
J.~N. Kather, J.~Krisam, P.~Charoentong, T.~Luedde, E.~Herpel, C.-A. Weis,
  T.~Gaiser, A.~Marx, N.~A. Valous, D.~Ferber \emph{et~al.}, ``Predicting
  survival from colorectal cancer histology slides using deep learning: A
  retrospective multicenter study,'' \emph{PLoS medicine}, vol.~16, no.~1, p.
  e1002730, 2019.

\bibitem{yang2021medmnist}
J.~Yang, R.~Shi, D.~Wei, Z.~Liu, L.~Zhao, B.~Ke, H.~Pfister, and B.~Ni,
  ``Medmnist v2: A large-scale lightweight benchmark for 2d and 3d biomedical
  image classification,'' \emph{arXiv preprint arXiv:2110.14795}, 2021.

\bibitem{macenko2009method}
M.~Macenko, M.~Niethammer, J.~S. Marron, D.~Borland, J.~T. Woosley, X.~Guan,
  C.~Schmitt, and N.~E. Thomas, ``A method for normalizing histology slides for
  quantitative analysis,'' in \emph{2009 IEEE international symposium on
  biomedical imaging: from nano to macro}.\hskip 1em plus 0.5em minus
  0.4em\relax IEEE, 2009, pp. 1107--1110.

\bibitem{wang2020covid}
L.~Wang, Z.~Q. Lin, and A.~Wong, ``Covid-net: A tailored deep convolutional
  neural network design for detection of covid-19 cases from chest x-ray
  images,'' \emph{Scientific Reports}, vol.~10, no.~1, pp. 1--12, 2020.

\bibitem{thabtah2017autism}
F.~Thabtah, ``Autism spectrum disorder screening: machine learning adaptation
  and dsm-5 fulfillment,'' in \emph{Proceedings of the 1st International
  Conference on Medical and health Informatics 2017}, 2017, pp. 1--6.

\bibitem{gardner1998artificial}
M.~W. Gardner and S.~Dorling, ``Artificial neural networks (the multilayer
  perceptron)—a review of applications in the atmospheric sciences,''
  \emph{Atmospheric environment}, vol.~32, no. 14-15, pp. 2627--2636, 1998.

\bibitem{o2015introduction}
K.~O'Shea and R.~Nash, ``An introduction to convolutional neural networks,''
  \emph{arXiv preprint arXiv:1511.08458}, 2015.

\bibitem{he2016deep}
K.~He, X.~Zhang, S.~Ren, and J.~Sun, ``Deep residual learning for image
  recognition,'' in \emph{Proceedings of the IEEE conference on computer vision
  and pattern recognition}, 2016, pp. 770--778.

\bibitem{leino2020stolen}
K.~Leino and M.~Fredrikson, ``Stolen memories: Leveraging model memorization
  for calibrated $\{$White-Box$\}$ membership inference,'' in \emph{29th USENIX
  security symposium (USENIX Security 20)}, 2020, pp. 1605--1622.

\bibitem{yeom2018privacy}
S.~Yeom, I.~Giacomelli, M.~Fredrikson, and S.~Jha, ``Privacy risk in machine
  learning: Analyzing the connection to overfitting,'' in \emph{2018 IEEE 31st
  computer security foundations symposium (CSF)}.\hskip 1em plus 0.5em minus
  0.4em\relax IEEE, 2018, pp. 268--282.

\bibitem{song2019privacy}
L.~Song, R.~Shokri, and P.~Mittal, ``Privacy risks of securing machine learning
  models against adversarial examples,'' in \emph{Proceedings of the 2019 ACM
  SIGSAC Conference on Computer and Communications Security}, 2019, pp.
  241--257.

\bibitem{shokri2017membership}
R.~Shokri, M.~Stronati, C.~Song, and V.~Shmatikov, ``Membership inference
  attacks against machine learning models,'' in \emph{2017 IEEE symposium on
  security and privacy (SP)}.\hskip 1em plus 0.5em minus 0.4em\relax IEEE,
  2017, pp. 3--18.

\bibitem{salem2018ml}
A.~Salem, Y.~Zhang, M.~Humbert, P.~Berrang, M.~Fritz, and M.~Backes,
  ``Ml-leaks: Model and data independent membership inference attacks and
  defenses on machine learning models,'' \emph{arXiv preprint
  arXiv:1806.01246}, 2018.

\bibitem{song2020systematic}
L.~Song and P.~Mittal, ``Systematic evaluation of privacy risks of machine
  learning models,'' \emph{arXiv preprint arXiv:2003.10595}, 2020.

\end{thebibliography}
}

\end{document}